\documentclass[10pt,twocolumn,letterpaper]{article}

\usepackage[pagenumbers]{cvpr} 

\usepackage{graphicx}
\usepackage{amsmath}
\usepackage{amssymb}
\usepackage{booktabs}
\usepackage{multirow}
\usepackage{mathrsfs}
\usepackage{appendix}

\usepackage[pagebackref,breaklinks,colorlinks]{hyperref}
\newcommand{\tabincell}[2]{\begin{tabular}{@{}#1@{}}#2\end{tabular}}

\usepackage[table]{xcolor}
\usepackage[capitalize]{cleveref}
\crefname{section}{Sec.}{Secs.}
\Crefname{section}{Section}{Sections}
\Crefname{table}{Table}{Tables}
\crefname{table}{Tab.}{Tabs.}

\newcommand{\yxq}[1]{{\color{green} }}

\newcommand{\ourmoudle}{progressive filtering strategy }
\def\cvprPaperID{xxx} 
\def\confName{CVPR}
\def\confYear{2023}

\begin{document}

\title{CrowdCLIP: Unsupervised Crowd Counting via Vision-Language Model}

\author{Dingkang Liang$^{*1}$, Jiahao Xie$^{*2}$, Zhikang Zou$^{3}$, Xiaoqing Ye$^{3}$, Wei Xu$^{2}$, Xiang Bai$^{\dag1}$\\
\\
        $^{1}$Huazhong University of Science and Technology, \{dkliang, xbai\}@hust.edu.cn\\
        $^{2}$Beijing University of Posts and Telecommunications, \{xiejiahao, xuwei2020\}@bupt.edu.cn\\
        $^{3}$ Baidu Inc., China\\
}

\maketitle
\protect \renewcommand{\thefootnote}{\fnsymbol{footnote}}
\footnotetext[1]{Equal contribution. $^{\dag}$Corresponding author.} 
\footnotetext[0]{Work done when Dingkang Liang was an intern at Baidu.}

\begin{abstract}

Supervised crowd counting relies heavily on costly manual labeling, which is difficult and expensive, especially in dense scenes. To alleviate the problem, we propose a novel unsupervised framework for crowd counting, named CrowdCLIP. The core idea is built on two observations: 1) the recent contrastive pre-trained vision-language model (CLIP) has presented impressive performance on various downstream tasks;
2) there is a natural mapping between crowd patches and count text. To the best of our knowledge, CrowdCLIP is the first to investigate the vision-language knowledge to solve the counting problem. Specifically, in the training stage, we exploit the multi-modal ranking loss by constructing ranking text prompts to match the size-sorted crowd patches to guide the image encoder learning. In the testing stage, to deal with the diversity of image patches, we propose a simple yet effective \ourmoudle to first select the highly potential crowd patches and then map them into the language space with various counting intervals. Extensive experiments on five challenging datasets demonstrate that the proposed CrowdCLIP achieves superior performance compared to previous unsupervised state-of-the-art counting methods. Notably, CrowdCLIP even surpasses some popular fully-supervised methods under the cross-dataset setting. The source code will be available at \url{https://github.com/dk-liang/CrowdCLIP}.

\end{abstract}

\section{Introduction}
\label{sec:intro}


Crowd counting aims to estimate the number of people from images or videos in various crowd scenes, which has received tremendous attention due to its wide applications in public safety and urban management~\cite{sindagi2018survey,kang2018beyond}. It is very challenging to accurately reason the count, especially in dense regions where the crowd gathers.

\begin{figure}[t]
	\begin{center}
		\includegraphics[width=0.96\linewidth]{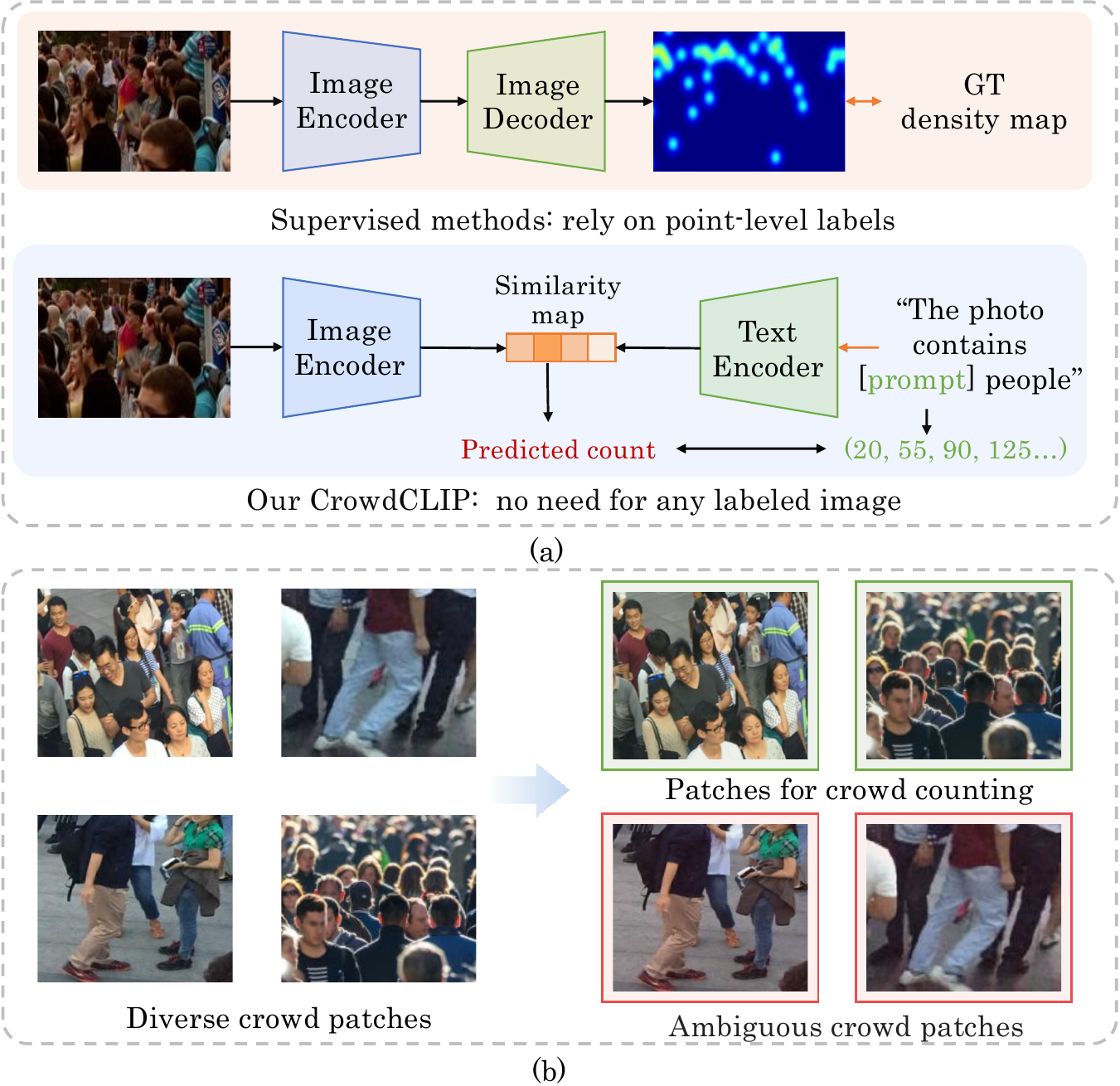}
	\end{center}
 \vspace{-10pt}
	\caption{
	(a) The supervised methods require point-level annotations, which need heavy manual labor to label a large-scale dataset. The proposed method transfers the vision-language knowledge to perform unsupervised crowd counting without any annotation; (b) Crowd counting aims to calculate the number of human heads, while some crowd patches do not contain human heads, \textit{i.e.}, ambiguous patches. }
	\label{fig:intro1}
\end{figure}
The recent crowd counting methods~\cite{sindagi2017cnn,zhang2016single,li2018csrnet,bai2020adaptive} attempt to regress a density map (Fig.~\ref{fig:intro1}(a)). To train such density-based models, point-level annotations are required, \textit{i.e.}, assigning a point in each human head. However, annotating point-level object annotations is an expensive and laborious process. For example, the NWPU-Crowd~\cite{gao2020nwpu} dataset, containing $5,109$ images, needs $30$ annotators and $3,000$ human hours for the entire annotation process. To reduce the annotation cost, some weakly-supervised methods~\cite{liang2022transcrowd,lei2021towards,yang2020weakly} and semi-supervised methods~\cite{liu2020semi,lin2022semi} are proposed, where the former usually adopts the count-level annotation as supervision, and the latter uses a small fraction of fully-labeled images and massive unlabeled images for training. However, both weakly and semi-supervised methods still need considerable label costs, especially when annotating dense or blurry images. 


Considering the above issues, a crowd counting model that can be trained without any labeled data is worth exploring. 
So far, there is only one approach called CSS-CCNN~\cite{babu2022completely} for pure unsupervised crowd counting. Based on the idea that natural crowds follow a power law distribution, CSS-CCNN ensures the distribution of predictions is matched to the prior. 
Though the performance of CSS-CCNN is better than the random paradigm, there is a significant performance gap compared to the popular fully supervised methods~\cite{zhang2016single,sam2017switching}. 
Recently, Contrastive Language-Image Pre-Training (CLIP) as a new paradigm has drawn increasing attention due to its powerful transfer ability. By using large-scale noisy image-text pairs to learn visual representation, CLIP  has achieved promising performance on various downstream vision tasks 
(\eg, object detection~\cite{shi2022proposalclip}, semantic segmentation~\cite{xu2022simple}, generation~\cite{hong2022avatarclip}). Whereas, how to apply such a language-driven model to crowd counting has not been explored. Obviously, CLIP cannot be directly 
applied to the counting task since there is no such count supervision during the contrastive pre-training of CLIP.

A natural way to exploit the vision-language knowledge is to discretize the crowd number into a set of intervals, which transfers the crowd counting to a classification instead of a regression task. Then one can directly calculate the similarity between the image embedding from the image encoder and the text embedding from the text encoder and choose the most similar image-text pair as the prediction count (called zero-shot CLIP). However, we reveal that the zero-shot CLIP reports unsatisfactory performance, attributed to two crucial reasons: 1) The zero-shot CLIP can not well understand crowd semantics since the original CLIP is mainly trained to recognize single-object images~\cite{shi2022proposalclip}; 2) Due to the non-uniform distribution of the crowd, the image patches are of high diversity while counting aims to calculate the number of human heads within each patch. Some crowd patches that do not contain human heads may cause ambiguity to CLIP, as shown in Fig.~\ref{fig:intro1}(b).


To relieve the above problems, in this paper, we propose CrowdCLIP, which adapts CLIP's strong vision-category correspondence capability to crowd counting in an unsupervised manner, as shown in Fig.~\ref{fig:intro1}(a). Specifically, first, we construct ranking text prompts to describe a set of size-sorted image patches during the training phase. As a result, the image encoder can be fine-tuned to better capture the crowd semantics through the multi-modal ranking loss. Second, during the testing phase, we propose a simple yet effective \ourmoudle consisting of three stages to choose high-related crowd patches. 
In particular, the first two stages aim to choose the high-related crowd patches with a coarse-to-fine classification paradigm, and the latest stage is utilized to map the corresponding crowd patches into an appropriate count. Thanks to such a progressive inference strategy, we can effectively reduce the impact of ambiguous crowd patches. 

Extensive experiments conducted on five challenging datasets in various data settings demonstrate the effectiveness of our method. In particular, our CrowdCLIP significantly outperforms the current unsupervised state-of-the-art method CSS-CCNN~\cite{babu2022completely} by $\mathbf{35.2}\%$ on the challenging UCF-QNRF dataset in terms of the MAE metric. 
Under cross-dataset validation, our method even surpasses some popular fully-supervised works~\cite{zhang2016single,shi2018crowd}.

Our major contributions can be summarized as follows: 1) In this paper, we propose a novel unsupervised crowd counting method named CrowdCLIP, which innovatively views crowd counting as an image-text matching problem. To the best of our knowledge, this is the first work to transfer vision-language knowledge to crowd counting. 2) We introduce a ranking-based contrastive fine-tuning strategy to make the image encoder better mine potential crowd semantics. In addition, a \ourmoudle is proposed to choose the high-related crowd patches for mapping to an appropriate count interval during the testing phase.

\section{Related Works}
\subsection{Fully-Supervised Crowd Counting}
The mainstream idea of supervised methods~\cite{sindagi2017cnn,zou2019attend,li2018csrnet,xu2019learn,zou2021coarse,liu2021exploiting,liu2021visdrone} is to regress a density map, which is generated from an elaborately labeled point map. In general, the labeled points are hard to reflect the size of the head, meaning the density-based paradigm easily meets the huge variation issue. To tackle the scale variations, various methods make many attempts. Specifically, some works~\cite{zhang2016single,sindagi2017cnn} adopt multi-column networks to learn multi-scale feature information. Some methods propose to utilize the scaling mechanism~\cite{jiang2020attention,sajid2020zoomcount,xu2022autoscale,xu2019learn} or scale selection~\cite{song2021choose} to relieve the scale variations. The attention mechanism is also a valuable tool to improve the feature representation, such as self-attention~\cite{lin2022boosting}, spatial attention~\cite{shi2019counting,xu2021dilated}, and other customized attention blocks~\cite{miao2020shallow,chen2021variational,xie2023super}. Different from regressing density maps, some methods~\cite{xiong2022discrete,wang2021uniformity,xiong2019open} leverage supervised-classifier to classify the crowd into different intervals, achieving appealing performance.

Another trend is based on localization~\cite{sam2020locate,wang2021self,liang2022focal,abousamra2020localization,wan2021generalized,wen2021detection,chen2021cell}, which can be divided into three categories: predict pseudo-bounding boxes~\cite{sam2020locate,liu2019point,wang2021self} or customize special localization-based maps~\cite{xu2022autoscale,liang2022focal,abousamra2020localization}, the other methods~\cite{liang2022end,song2021rethinking} directly regress the point coordinates, removing the need for pre-processing or post-processing.



\subsection{Weakly-/Semi-/Unsupervised Crowd Counting}

The fully-supervised methods need expensive costs to label points for each head. To this end, weakly or semi-supervised methods are proposed to reduce the annotations burden. The weakly-supervised methods~\cite{lei2021towards,yang2020weakly,kong2020weakly,liang2022transcrowd} suggest using count-level instead of point-level annotation as the supervision. The semi-supervised methods~\cite{liu2020semi,meng2021spatial,xu2021crowd} leverage small-label data to train a model and further use massive unlabeled data to improve the performance. Method in~\cite{sam2019almost} optimizes almost 99\% of the model parameters with unlabeled data. However, all the above methods still require some annotated data. Once the labeled data is left, these models can not be trained.


So far, only one method, CSS-CCNN~\cite{babu2022completely}, focuses on pure unsupervised settings, \ie, training a model without any label. CSS-CCNN~\cite{babu2022completely} argues that natural crowds follow a power law distribution, which could be leveraged to yield error signals for back-propagation. We empirically find that there is a significant performance gap between CSS-CCNN and some popular fully-supervised methods~\cite{zhang2016single,zhang2015cross}. In this paper, we propose a novel method named CrowdCLIP, which transfers the counting into an image-text matching problem, to boost the performance for unsupervised crowd counting by a large margin.



\subsection{Vision-Language Contrastive Learning}

Recently, vision-language pre-training (CLIP~\cite{radford2021learning}) using massive image-text pairs from the Internet has attracted more and more attention. Several methods~\cite{shi2022proposalclip,xu2022simple,pakhomov2021segmentation,hong2022avatarclip,liordinalclip} transfer the vision-language correspondence to the downstream tasks, such as object detection~\cite{shi2022proposalclip}, semantic segmentation~\cite{xu2022simple}, and generation\cite{hong2022avatarclip}. Benefiting from the strong zero-shot ability of CLIP, these unsupervised methods achieve promising performance. In this paper, we study how to transfer the vision-language knowledge to the unsupervised crowd counting task. 

\begin{figure*}[t]
	\begin{center}
		\includegraphics[width=0.96\linewidth]{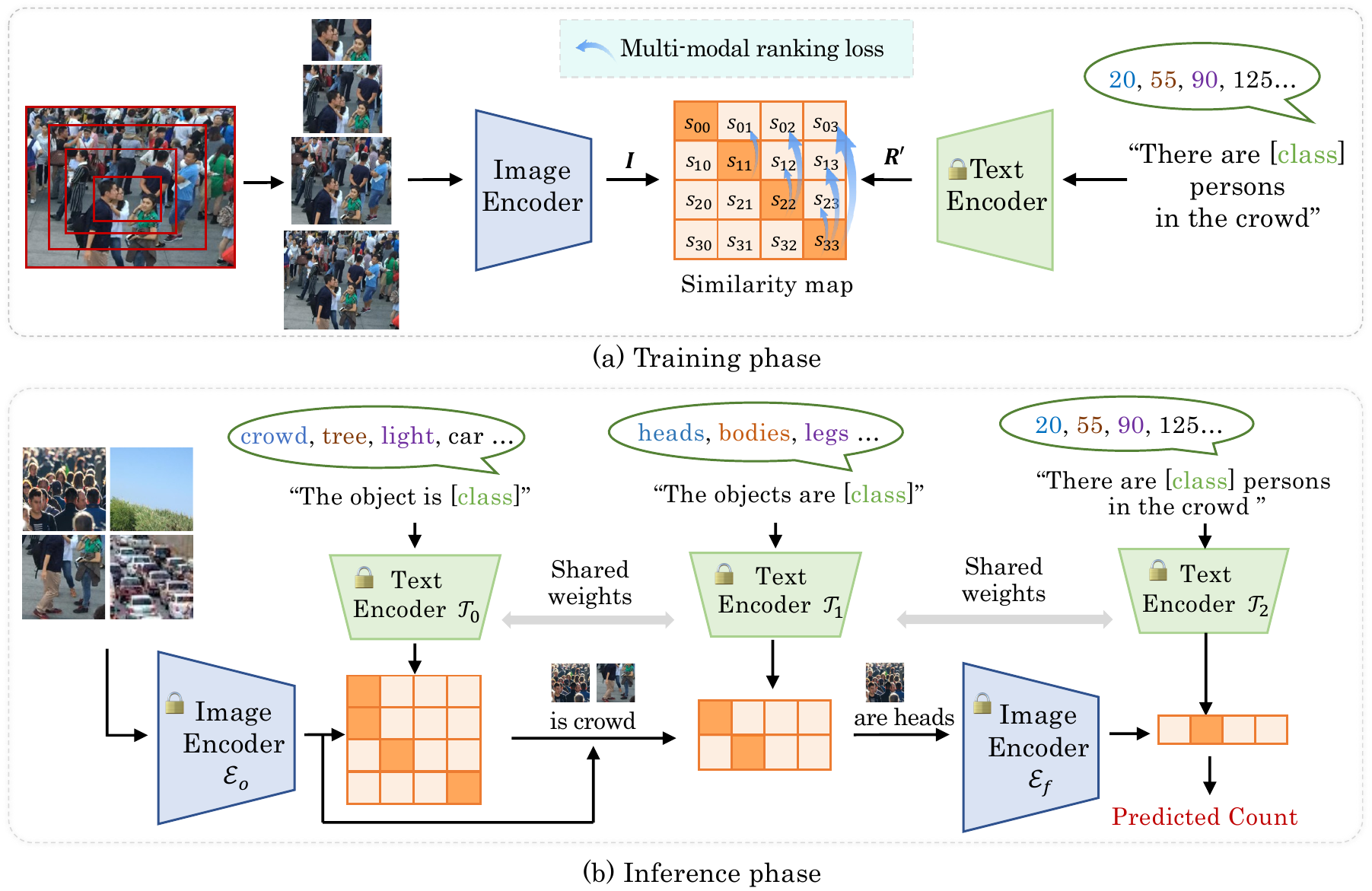}
	\end{center}
 	\vspace{-15pt}
	\caption{Overview of our framework. (a) During the training phase, we fine-tune the image encoder by introducing ranking prompts while the parameters of the text encoder are frozen. (b) During the testing phase, we propose a \ourmoudle that progressively queries the most likely crowd patches and maps the filtered patches into the specific quantitive count. }
	\label{fig:pipeline}
\end{figure*}

\section{Preliminary}
In this section, we revisit CLIP~\cite{radford2021learning} and then introduce the setting of unsupervised crowd counting.
\subsection{A Revisit of CLIP}
CLIP~\cite{radford2021learning}, a representative pre-trained vision-language model, focuses on how to build the connection between visual concepts and language concepts. CLIP contains two encoders used for encoding the image feature and text feature, respectively. Given an image-text pair, the CLIP aims to compute the semantic similarity between the encoding image feature and the encoding text feature. The pre-trained CLIP can be easily extended to the zero-shot/open-vocabulary image classification. Specifically, one can utilize a series of class names (\eg, `cat,' `dog') for replacing the pre-defined text prompt template, \eg, ``a photo of [CLASS]". Then the text is fed into the text encoder to generate the class embeddings used to compute the similarity with image embedding for classification. 

In this paper, we are the first to study how to extend the strong correspondence between the image and text of CLIP to the crowd counting task.

\subsection{Unsupervised Crowd Counting}
The goal of crowd counting is to estimate the pedestrian number given crowd images. Following the unsupervised definition from previous methods~\cite{doersch2015unsupervised,babu2022completely}, \ie, during the training, the model does not need a single annotated image while allowing to use of the supervision provided by the original data. The manually annotated validation or test set is only used for evaluation. Note that using CLIP~\cite{radford2021learning} to the downstream tasks without the task-related training label is in an accepted unsupervised manner~\cite{shi2022proposalclip,abdal2022clip2stylegan,zhou2022extract}. 


\section{Our Method}

The overview of our method is shown in Fig.~\ref{fig:pipeline}. During the training phase, we fine-tune the image encoder by introducing ranking text prompts while the parameters of the text encoder are frozen. During the testing phase, we propose a \ourmoudle that progressively queries the highly potential crowd patches and maps the filtered patches into the specific crowd intervals.

\subsection{Ranking-based Contrastive Fine-tuning}
In this section, we introduce how to fine-tune the CLIP~\cite{radford2021learning} to improve the ability to extract crowd semantics. Note that the original CLIP needs image-text pairs to complete the contrastive pre-trained process. However, no label is provided in the unsupervised setting, \ie, the fine-tuning lacks the corresponding text modality as supervision. To this end, we construct ranking prompts using texts of counting intervals to describe the size-ordered input images, as shown in Fig.~\ref{fig:pipeline}(a). As a result, the image encoder can be fine-tuned through multi-modal ranking loss.

\noindent\textbf{Image to patches.} Given an input crowd image, we first crop a set of square patches $\{\mathcal{O}_{M}\}$, where $M$ is the predefined number of patches. The cropped patches obey the following rules:
\begin{itemize}
    \item For any two cropped image patches ($\mathcal{O}_{i}$ and $\mathcal{O}_{j}$, $0 \leq i<j \leq M-1$), the size of patch $\mathcal{O}_{i}$ is smaller than the size of patch $\mathcal{O}_{j}$. 
    \item The patches from the same image share the same image center. During the training, all patches are resized to the same size and fed into the image encoder to generate the image rank embeddings $\mathbf{I} = [\mathbf{I}_0, \mathbf{I}_1,..., \mathbf{I}_{M-1}]$, where $\mathbf{I} \in \mathbb{R}^{M \times C}$. 
\end{itemize}

\noindent\textbf{Prompt design.}
The original CLIP~\cite{radford2021learning} is not designed for the counting task. How to customize feasible text prompts needs to be studied. As mentioned above, we have successfully collected a series of patches whose sizes are ordinal. Obviously, the number of human heads from different patches is ordinality, and the larger patches correspond to the more or equal number of human heads. Thus, we design ranking text prompts to describe the ordinal relationship of image patches. Specifically, we propose to learn the rank embeddings to preserve the order of the image patches in the language latent space. The text prompt is defined as ``There are [class] persons in the crowd", where [class] represents a set of base rank number $R = [R_0, R_0 + K, ..., R_0 + (N-1) K]$, where $R_0$, $K$ and $N$ denote the basic reference count, counting interval and number of class, respectively. 
The text prompts will be fed into the text encoder to obtain the output text rank embeddings $\mathbf{R}' = [\mathbf{R}'_0, \mathbf{R}'_1,..., \mathbf{R}'_{N-1}]$, where $\mathbf{R}' \in \mathbb{R}^{N \times C}$. 

\noindent\textbf{Image encoder optimization.}
Suppose we have a set of image rank embeddings $\mathbf{I} \in \mathbb{R}^{M \times C}$ and text rank embeddings $\mathbf{R}' \in \mathbb{R}^{N \times C}$, where $\mathbf{I}$ and $\mathbf{R}'$ are obtained from the image encoder and text encoder, respectively. For the image-language matching pipeline, we
calculate the similarity scores between $\mathbf{I}$ and $\mathbf{R}'$ via inner product and obtain the similarity matrix $\mathbf{S} = [\mathbf{s}_{izj}]$, where $\mathbf{S} \in \mathbb{R}^{M \times N}$:
\begin{equation}
   \mathbf{s}_{i,j} = \mathbf{I}_i \cdot {\mathbf{R}'}_j^T, 
\end{equation}
where $\mathbf{I}_i \in \mathbb{R}^{1 \times C}$, $\mathbf{R}'_j \in \mathbb{R}^{1 \times C}$, $0 \leq i \leq M - 1$ and $0 \leq j \leq N - 1$. Due to the inheritance of ranking relationships from the images and text prompts, we hope the similarity matrix $\mathbf{S}$ is a specific ordinal matrix (Fig.~\ref{fig:pipeline}(a)):
\begin{equation}
  \mathbf{s}_{i',i} \leq \mathbf{s}_{i, i},
  \label{eq2}
\end{equation}
where $ 0 \leq i' \leq i \leq M - 1$. 
To preserve the order of the image-text pair in the latent space, we propose to optimize the image encoder through the multi-modal ranking loss. Specifically, we use the principal diagonal of the similarity matrix as the base and calculate the ranking loss from the bottom up:
\begin{equation}
  L_r = \max(0, \mathbf{s}_{i', i}- \mathbf{s}_{i,i}).
  \label{eq3}
\end{equation}

We set $M = N$ in practice to guarantee that $\mathbf{S}$ is a square matrix. Eq.2 and Eq.3 build intrinsic correspondence between size-sorted patches and ranking text prompts, resulting in the similarity matrix (optimization goal, similar to multi-modal similarity matrix like PointCLIP~\cite{zhang2022pointclip}, AudioCLIP~\cite{guzhov2022audioclip}).
During the fine-tuning, the weights of the text encoder are frozen, \ie, $L_r$ aims to align the image embedding into the fixed ranking language space. In this way, the text embeddings are constrained in the well-learned language latent space, leading to robust generalization. 

Note that the ranking loss we used is different from the previous methods~\cite{liu2018leveraging,liu2019exploiting}. First, they calculate the ranking loss from one modality (\ie, only images), while ours is designed for the multi-modal (image and text). Second, they still demand labeled data since the ranking loss they used just can judge the order of given image patches, which cannot directly predict the number range of patches. In contrast, our approach does not need any labeled crowd images.

\subsection{Progressive Filtering Strategy}
In this part, we introduce the detail of the proposed progress filtering strategy consisting of three stages,  used to select the real crowd patches and map them into appropriate count intervals at the inference stage, as shown in Fig.~\ref{fig:pipeline}(b). 
For simplicity, we name the original image encoder and fine-tuned image encoder as $\mathcal{E}_o$ and $\mathcal{E}_f$, respectively. $\mathcal{E}_o$ is used to choose the high-confidence crowd patches, and $\mathcal{E}_f$ is used for the final counting. 

Given an input image, we first divide it into a grid of $P \times P$ patches, then the patches and corresponding text prompts will be respectively fed into $\mathcal{E}_o$ and the first text encoder $\mathcal{T}_0$ to generate similarity scores for coarse classification. The text prompt of $\mathcal{T}_0$ is set to ``The object is [class]", which aims to classify the patches into different categories with clear distinction (\eg, `crowd', `tree', `car'). 

The selected crowd patches from the first stage may contain different components of the human, such as human heads, bodies, and legs. However, the crowd counting task aims to estimate the number of human heads instead of other components since only the heads are not easily obscured compared with other components. Thus, in the second stage, we adopt fine-grain text prompts and feed them into the second text encoder $\mathcal{T}_1$ to further filter the patches. The fine-grain text prompts are defined as ``The objects are [class]," where [class] are some fine-grained categories (\eg, `human heads,' `human bodies'). As a result, high-confidence crowd patches with human heads are obtained. Note that in this stage, we still use the non-fine-tuned image encoder $\mathcal{E}_o$ for calculating the similarity.

In the third stage, we adopt the fine-tuned $\mathcal{E}_f$ as the image encoder, and the text prompts of the third $\mathcal{T}_2$ are the same as the fine-tuning phase, \ie, the ranking text prompts are defined as ``There are [class] persons in the crowd," where [class] is the pre-defined ranking number $R$. The final count can be obtained by choosing the most similar image-text pair based on the image embedding and class embeddings from $\mathcal{E}_f$ and $\mathcal{T}_2$.

Note that in practice, $\mathcal{T}_{0}$, $\mathcal{T}_{1}$ and $\mathcal{T}_{2}$ share the same parameters, and all images share the same text prompts. In other words, we can get the text embeddings in advance instead of processing the text prompts in each inference phase, thus keeping efficiency. 

\begin{table*}[t]
\centering
\footnotesize
\caption{Comparison of the counting performance on the UCF-QNRF, JHU-Crowd++, ShanghaiTech Part A, Part B, and UCF\_CC\_50 datasets. Random* denotes that we randomly select a value from the pre-defined rank number $R$ for each cropped patch. 
}

\label{tab:main_counting}
\setlength{\tabcolsep}{2.0mm}{
\begin{tabular}{lcccccccccccc}
 \toprule
 {\multirow{3}{*}{Method}} & {\multirow{3}{*}{Year}} & {\multirow{3}{*}{\tabincell{c}{Label}}} &\multicolumn{2}{c}{\multirow{2}*{QNRF}} &\multicolumn{2}{c}{\multirow{2}*{JHU}} &\multicolumn{2}{c}{\multirow{2}*{Part A}} &\multicolumn{2}{c}{\multirow{2}*{Part B}} &\multicolumn{2}{c}{\multirow{2}*{UCF\_CC\_50}}\\
&&&&&&&&&\\
\cmidrule(r){4-5} \cmidrule(r){6-7} \cmidrule(r){8-9} \cmidrule(r){10-11} \cmidrule(r){12-13}   
&&& MAE & MSE& MAE & MSE &MAE&MSE &MAE&MSE &MAE&MSE\\
\midrule
Zhang \textit{et al.}~\cite{zhang2015cross} & CVPR 15 & Point &-&-&-&-&181.8&277.7&32.0&49.8&467.0&498.5\\
MCNN~\cite{zhang2016single} &CVPR 16 & Point &277.0&426.0 &188.9&483.4 &110.2&173.2 & 26.4 & 41.3 &377.6&509.1 \\
Switch CNN~\cite{sam2017switching} & CVPR 17 & Point & 228.0 &445.0 &-&-& 90.4 & 135.0 & 21.6 &33.4 &318.1&439.2\\
LSC-CNN~\cite{sam2020locate} & TPAMI 21 & Point & 120.5&218.2&112.7&454.4& 66.4&117.0&8.1&12.7 & 225.6&302.7\\
CLTR~\cite{liang2022end}&ECCV 22&Point &85.8&141.3 &59.5&240.6&56.9&95.2&6.5&10.2 &-&-\\
\midrule
CSS-CCNN-Random~\cite{babu2022completely} &ECCV 22&\textcolor{red}{None} &718.7 &1036.3 & 320.3 &793.5  &431.1 &559.0&-&-&1279.3&1567.9\\
Random* &-&\textcolor{red}{None} &633.6 &978.9 & 297.5 &801.6  &411.5 &511.1&158.7&287.4&1251.6&1497.8\\
CSS-CCNN~\cite{babu2022completely} & ECCV 22&\textcolor{red}{None} &437.0 &722.3 & 217.6 &651.3  &197.3 &295.9&-&-&564.9&959.4\\
CrowdCLIP (\textbf{ours}) &-&\textcolor{red}{None} &\textbf{283.3} &\textbf{488.7} & \textbf{213.7} &\textbf{576.1} &\textbf{146.1} &\textbf{236.3}&\textbf{69.3}&\textbf{85.8}&\textbf{438.3}&\textbf{604.7}\\
\rowcolor{green!12}\textit{Improvement} &-&- & \textit{35.2\%} &\textit{32.3\%} &\textit{1.8\%} & \textit{11.5\%} &\textit{26.0\%} & \textit{20.1\%}&\textit{56.3\%}&\textit{70.1\%}&\textit{22.4\%} &\textit{37.0\%} \\
     \bottomrule
\end{tabular}}
\end{table*}

\begin{figure*}[t]
	\begin{center}
		\includegraphics[width=0.96\linewidth]{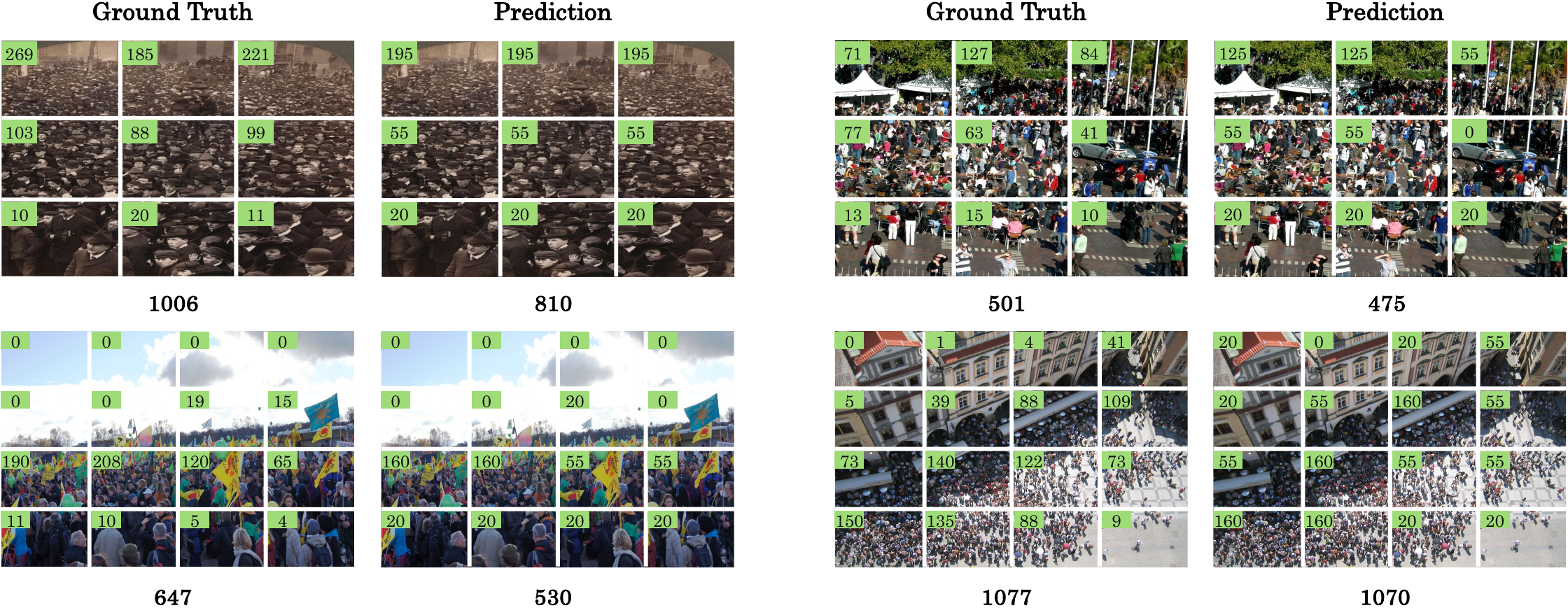}
	\end{center}
	\vspace{-15pt}
	\caption{The first and second rows are selected from the ShanghaiTech Part A and UCF-QNRF datasets, respectively. }
	\label{fig:visualization}
\end{figure*}

\begin{table*}[t]
\footnotesize
\centering
\setlength{\tabcolsep}{6.5mm}
\caption{
Experimental results on the transferability of our method and popular fully-supervised methods under cross-dataset evaluation. }
\resizebox{0.95\linewidth}{!}{
\begin{tabular}{ lcccccccc }
    \toprule
    {\multirow{2}{*}{Method}}&{\multirow{2}{*}{Year}}&{\multirow{2}{*}{Label}}&\multicolumn{2}{c}{Part B$\rightarrow$Part A}&\multicolumn{2}{c}{Part A$\rightarrow$Part B}\\
    \cmidrule{4-7}
     &&& MAE & MSE&MAE&MSE\\
    \midrule
    MCNN~\cite{zhang2016single} &CVPR 16&Point & 221.4&357.8 & 85.2&142.3\\
    D-ConvNet~\cite{shi2018crowd}&ECCV 18&Point & \textbf{140.4}&\textbf{226.1}&\textbf{49.1}&99.2\\
    \midrule
    \textbf{CrowdCLIP (ours)} &-&\textcolor{red}{None}&217.0&322.7&69.6&\textbf{80.7}\\
    \bottomrule
    \end{tabular}
    }
	\label{tab:transfer_ab}%
\end{table*}

\begin{table*}[t]
\footnotesize
\centering
\setlength{\tabcolsep}{3.5mm}
\caption{Experimental results on the transferability of unsupervised methods under cross-dataset evaluation.  }
\resizebox{0.95\linewidth}{!}{
\begin{tabular}{ lccccccccccc }
    \toprule
    {\multirow{2}{*}{Method}}&{\multirow{2}{*}{Year}}&{\multirow{2}{*}{Label}}&\multicolumn{2}{c}{Part A$\rightarrow$QNRF}&\multicolumn{2}{c}{QNRF$\rightarrow$Part A}&\multicolumn{2}{c}{Part A$\rightarrow$JHU}
    &\multicolumn{2}{c}{JHU$\rightarrow$Part A}
    \\
    \cmidrule{4-11}
     &&& MAE & MSE&MAE&MSE&MAE&MSE&MAE&MSE\\
    \midrule
    CSS-CCNN~\cite{babu2022completely} &ECCV 22 &\textcolor{red}{None}&472.4&-&235.7&-&251.3&- &266.3&-\\
    \textbf{CrowdCLIP (ours)} &-&\textcolor{red}{None}&\textbf{294.9}&\textbf{498.7}&\textbf{148.2}&\textbf{227.3}&\textbf{212.8}&\textbf{508.5}&\textbf{253.2}&\textbf{393.2}\\
    \rowcolor{green!12} \textit{Improvement} &-&-&\textit{37.6\%}&-&\textit{37.1\%}&-&\textit{15.3\%}&-&\textit{4.9\%}&-\\
    \bottomrule
    \end{tabular}
}
	\label{tab:transfer_unsupervised}%

\end{table*}

\begin{table*}[t]
\footnotesize
\setlength{\tabcolsep}{6mm}
\centering
\caption{The effectiveness of our \ourmoudle and fine-tuning. $\mathcal{E}_o$ and $\mathcal{E}_f$ refer to the original image encoder and fine-tuned image encoder, respectively. $\mathcal{T}_o$, $\mathcal{T}_1$, and $\mathcal{T}_2$ denote the text encoder of the first, second and third text encoder. Random*  denotes that we randomly select a value from the pre-defined set of rank numbers $R$ for each cropped patch. }
\label{tab:effectiveness_fine-tune}
\resizebox{0.9\linewidth}{!}{
\begin{tabular}{ ccccccc }
\toprule

Methods & First stage & Second stage & Third stage  & MAE& MSE \\
\midrule
Random* &-&-&-&633.6 & 978.9\\
CSS-CCNN~\cite{babu2022completely} &-&-&-&437.0&722.3\\
\midrule
Zero-Shot CLIP~\cite{radford2021learning}& -  & - & $\mathcal{E}_o, \mathcal{T}_2$  &528.7 &690.7 \\
CrowdCLIP & -  & - & $\mathcal{E}_f, \mathcal{T}_2$& 318.8&491.9 \\
\midrule
Zero-Shot CLIP~\cite{radford2021learning}& $\mathcal{E}_o, \mathcal{T}_0$ &-&$\mathcal{E}_o, \mathcal{T}_2$ &437.7 & 623.1\\
CrowdCLIP & $\mathcal{E}_o, \mathcal{T}_0$  & - & $\mathcal{E}_f, \mathcal{T}_2$ & 286.8&490.4 \\
\midrule
Zero-Shot CLIP~\cite{radford2021learning}&$\mathcal{E}_o, \mathcal{T}_0$&$\mathcal{E}_o, \mathcal{T}_1$&$\mathcal{E}_o, \mathcal{T}_2$ &414.7&612.4\\
CrowdCLIP (\textbf{ours}) & $\mathcal{E}_o, \mathcal{T}_0$  & $\mathcal{E}_o, \mathcal{T}_1$ & $\mathcal{E}_f, \mathcal{T}_2$ & \textbf{283.3}&\textbf{488.7} \\
\bottomrule
\end{tabular}
}
\end{table*}

\begin{table}[t]
\footnotesize
\setlength{\tabcolsep}{3.0mm}
\centering
\caption{Ablation study on ranking prompts design (including the basic reference count $R_0$ and counting interval $K$). The ranking prompts are defined as ``There are [class] persons in the crowd."}
\label{tab:prompts}
\resizebox{0.94\linewidth}{!}{
\begin{tabular}{ ccccc }
\toprule

$R_0 $& $K$ &Prompts &MAE& MSE \\
\midrule
20 & 30 & $[$20, 50, ..., 140, 170$]$ &324.1&569.8\\
20 & 35 & $[$20, 55, ..., 160, 195$]$ &\textbf{283.3}&\textbf{488.7}\\
20 & 40 & $[$20, 60, ..., 180, 220$]$ &358.4&602.9\\
A + 20 & 35 & $[$A + 20, ..., A + 195$]$ &316.2&515.3\\ 
\midrule

10 & 35 & $[$10, 45, ..., 150, 185$]$ &374.5 &602.5\\
30 & 35 & $[$30, 65, ..., 170, 205$]$ &373.4 &633.7\\
\bottomrule
\end{tabular}
}
\end{table}

\section{Experiments}
\label{experiments}
\subsection{Dataset and Evaluation Metric}

\noindent\textbf{UCF-QNRF~\cite{idrees2018composition}} is a dense counting dataset. There are $1,535$ images with crowd numbers varying from $49$ to $12,865$. The images are split into the training set with $1,201$ images and the testing set with $334$ images.

\noindent\textbf{JHU-Crowd++~\cite{sindagi2020jhu-crowd++}} is one of the largest counting datasets. It contains $4,372$ images with $1,515,005$ annotations. Specifically, $2,272$, $500$, and $1,600$ images are divided into the training, validation, and testing sets. There is a large percentage of images captured on rainy and foggy days. These degraded images increase challenges for crowd counting.

\noindent\textbf{ShanghaiTech~\cite{zhang2016single}} includes Part A and Part B. Part A contains $482$ images with $241,677$ annotations, the training and testing sets consisting of $300$ and $182$ images, respectively. Part B contains $716$ images and a total of $88,488$ annotated head centers, where $400$ images are used for training and the rest $316$ images for testing.

\noindent\textbf{UCF-CC50~\cite{idrees2013multi}} is a challenging counting dataset. It contains only $50$ images but has $63,075$ annotated individuals, where the crowd numbers vary from $94$ to $4,543$.

\noindent\textbf{Evaluation metric.} We use the mean absolute error (MAE) and mean square error (MSE) to evaluate the crowd counting performance. The two counting metrics are defined as:
$MAE=\frac{1}{N_{c} } {\textstyle \sum_{i=1}^{N_{c}}}\left | E_{i}-C_{i} \right |$,
$MSE=\sqrt{\frac{1}{N_{c} }\textstyle \sum_{i=1}^{N_{c}}\left | E_{i}-C_{i} \right |^{2} }$,
where $N_{c}$ is the number of images in the test set. $E_{i}$ and $C_{i}$ represent the estimated count and the ground truth of the $i$-th image, respectively.

\subsection{Implement Details}

All experiments are conducted on an Nvidia 3090 GPU. CLIP with ViT-B/16 backbone is used. In the fine-tuning phase, the parameters of the text encoder are frozen, and the RAdam optimizer~\cite{liu2019variance} with a learning rate of 1e-4 is used to optimize the image encoder. The number of training epochs is set to $100$. The $M$ and $N$ are set to $6$. For the ranking text prompts, we set the basic reference count $R_0$ as $20$ and the counting interval $K$ as $35$, \ie, the text prompts are ``There are [20/55/90/125/160/195] persons in the crowd." Note that the ranking text prompts we used are kept the same in all datasets. During the testing phase, we set the $P$ as $4$ for UCF-QNRF and UCF\_CC\_50 datasets and set $P$ as $3$ for the rest datasets. For the large-scale datasets (\ie, UCF-QNRF, JHU-Crowd++), we make the longer size less than $2048$, keeping the original aspect ratio.

\section{Results and Analysis}
\subsection{Comparison with the State-of-the-Arts}


As reported in Tab.~\ref{tab:main_counting}, the proposed CrowdCLIP goes beyond the state-of-the-art method~\cite{babu2022completely} by a large margin in all evaluated datasets. Typically, for the UCF-QNRF, an extremely dense dataset, CrowdCLIP outperforms CSS-CCNN by $35.2\%$ improvement for MAE and $32.3$\% improvement for MSE. 
For the degraded images (JHU-Crowd++ dataset), our method achieves 213.7 MAE and 576.1 MSE. Our method also reports remarkable performance on two sparse datasets, ShanghaiTech Part A and ShanghaiTech Part B. These impressive results verify that our method is robust in various complex conditions. We further provide some qualitative visualizations to analyze the effectiveness of our method, as shown in Fig.~\ref{fig:visualization}. CrowdCLIP performs well in different scale scenes, although it is a pure unsupervised method. 

Additionally, we can find that our unsupervised method still presents highly competitive performance compared with some popular fully-supervised methods~\cite{zhang2016single,zhang2015cross}. Specifically, on the UCF-QNRF dataset, our method is very close to the MCNN~\cite{zhang2016single} in terms of MAE ($283.3$ vs. $277.0$). An interesting phenomenon is that our method achieves better performance than the method in ~\cite{zhang2015cross} on the ShanghaiTech Part A and UCF\_CC\_50 datasets.

\subsection{Cross-Dataset Validation }
Generally, scene variation could easily cause significant performance drops, while a crowd counting method with strong generalization ability is usually expected. So we adopt cross-dataset evaluation to demonstrate the generalization ability of CrowdCLIP. In this setting, the model is trained on one dataset while testing on the other.

\noindent\textbf{Compared with fully-supervised methods}. We first make comparisons between our CrowdCLIP and two fully-supervised methods~\cite{zhang2016single,shi2018crowd}, as depicted in Tab.~\ref{tab:transfer_ab}. Although our method does not require any annotated label, it still achieves competitive transfer abilities. Our method even outperforms MCNN~\cite{zhang2016single} by $18.3\%$ in MAE and outperforms D-ConvNet~\cite{shi2018crowd} by $18.6\%$ in MSE when adapting ShanghaiTech Part A to Part B.

\noindent\textbf{Compared with unsupervised methods}, our method significantly outperforms the state-of-the-art~\cite{babu2022completely} on various transfer settings, as shown in Tab.~\ref{tab:transfer_unsupervised}. Specifically, CrowdCLIP significantly improves the CSS-CCNN by $37.6\%$ MAE on Part A crossing to UCF-QNRF and $37.1\%$ MAE on UCF-QNRF crossing Part A. For the rest crossing setting, CrowdCLIP also achieves the best results.

\subsection{Ablation Study}

\label{ablation}
The following experiments are conducted on the UCF-QNRF~\cite{idrees2018composition} dataset, a large and dense dataset that can effectively avoid overfitting.

\noindent\textbf{The effectiveness of the \ourmoudle  and fine-tuning.} 
We first study the effectiveness of the proposed \ourmoudle and fine-tuning, and the results are listed in Tab.~\ref{tab:effectiveness_fine-tune}. We can make the following observations: 1) As expected, directly using the original CLIP (\ie, zero-shot CLIP without the first two text encoders) reports unsatisfactory performance, only $528.7$ MAE, significantly worse than the current SOTA~\cite{babu2022completely} ($437.0$ MAE). This result also verifies that the original CLIP can not directly work well in the crowd counting task, \ie, it is non-trivial to apply CLIP for our task; 2) Using the proposed inference strategy, we improve the performance of the original CLIP from $528.7$ MAE to $414.7$ MAE, which is slightly better than CSS-CCNN. 3) When we adopt the fine-tuned image encoder $\mathcal{E}_f$ for the last stage to map the crowd patches into count intervals, we get a significant performance gain. This highlights the effectiveness of ranking-based contrastive fine-tuning. The following ablation studies are organized using the setting of the last line of Tab.~\ref{tab:effectiveness_fine-tune}.

\begin{table}[t]
\footnotesize
\setlength{\tabcolsep}{2.0mm}
\centering
\caption{The influence of fixing different encoders.  }
\label{tab:fixed}
\resizebox{0.94\linewidth}{!}{
\begin{tabular}{ cccc }
\toprule
Fixed text encoder& Fixed image encoder&MAE & MSE \\
\midrule
\checkmark&\checkmark&414.7 &612.4\\
\checkmark&-&\textbf{283.3}&\textbf{488.7}\\
-&\checkmark&523.4 &870.4\\
-&-&416.6 &723.5\\
\bottomrule
\end{tabular}
}
\end{table}

 \begin{table}[t]
\footnotesize
\setlength{\tabcolsep}{5mm}
\centering
\caption{Ablation study on the patch number design.}
\label{tab:patch_size}
\resizebox{0.94\linewidth}{!}{
\begin{tabular}{ cccc }
\toprule
Setting &  Patch Number &MAE& MSE \\
\midrule
I & 3 & 325.0 & 534.6\\
II & 4 &\textbf{283.3} & 488.7\\
III & 5 &305.2&\textbf{477.8}\\
\bottomrule
\end{tabular}
}
\end{table}

\noindent\textbf{Analysis on ranking prompts design.}
We then analyze the influence of different ranking text prompt designs, as shown in Tab.~\ref{tab:prompts}. We observe that when the ranking prompts are set to [`20', `55', `90', `125', `160', `195'], \ie, $R_0 = 20$ and $K = 35$, the CrowdCLIP achieves the best performance. 
An interesting phenomenon is that when we set an abstract ranking prompt $R_0 = A + 20$ and  $K = 35$, where `A' is the original Latin alphabet, the CrowdCLIP also achieves superior performance. It indicates that CrowdCLIP can study the potential ranking representation from the ordinal language space.
Using different ranking prompts to fine-tune the image encoder, the performance is always better than the zero-shot CLIP~\cite{radford2021learning} and CSS-CCNN~\cite{babu2022completely}, which demonstrates the effectiveness of our CrowdCLIP. The following ablation studies are organized using $R_0 = 20$ and $K = 35$. 
 

\noindent\textbf{Analysis on fixing different encoders.} We further study the effect of fixing different encoders of CLIP~\cite{radford2021learning}, listed in Tab.~\ref{tab:fixed}. If we directly use the original CLIP with the proposed inference strategy (\ie, the parameters of the image encoder and text encoder are not tuned), the MAE is only $414.7$. When we only fine-tune the image encoder, we observe a significant improvement, where MAE is from $414.7$ to $283.3$. However, if we try to fine-tune the text encoder, we observe a significant performance drop. We argue the main reasons are as follows: 1) The natural language knowledge learned from large-scale pre-training encodes rich language priors, and fine-tuning will break the language priors. 2) As the text encoder is fixed, mapping image features into ranking language space will have a strong reference. 

\noindent\textbf{The influence of patch number.} We next study the influence of using different patch number $P$ during the testing phase. As shown in Tab.~\ref{tab:patch_size}, the MAE and MSE achieve the best when the $P$ is set to $4$ and $5$, respectively. Note that whether it is set to $3$, $4$ or $5$, the performance of CrowdCLIP always significantly outperforms the zero-shot CLIP~\cite{radford2021learning}.

\begin{figure}[t]
	\begin{center}
		\includegraphics[width=0.96\linewidth]{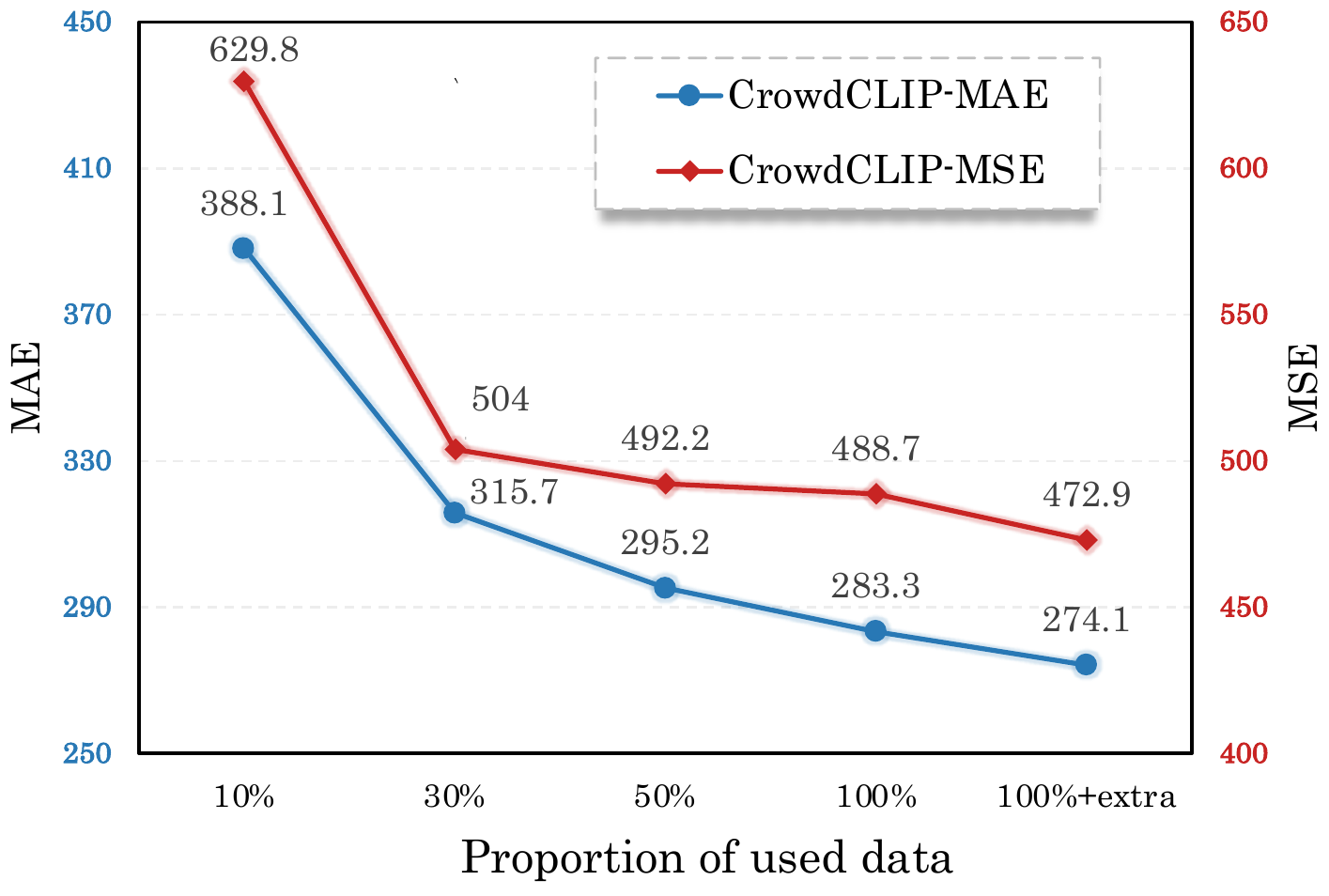}
	\end{center}
	\vspace{-10pt}
	\caption{Comparison in MAE and MSE of using different data size for training on UCF-QNRF dataset. We use ShanghaiTech Part A as the extra data.}
	\label{fig:extra_data}
\end{figure}

 \noindent\textbf{The influence of data size.} Finally, we explore the influence of using different data sizes. Fig.~\ref{fig:extra_data} shows that performance improves steadily as the data number increases.
Additionally, since our method is in an unsupervised setting, we can further utilize extra data to improve the performance. For example, when adopting ShanghaiTech Part A as the extra data, the proposed CrowdCLIP achieves considerable performance gain. These impressive results emphasize the practical utility of our CrowdCLIP.

\subsection{Limitation} 

The main limitation is that our method only provides the count-level information for a given image. However, point-level information is also useful to help analyze the crowd. In the future, we would like to explore localization in an unsupervised manner for crowd counting.

\section{Conclusion}
In this work, we propose a new framework, CrowdCLIP, to transfer the knowledge from the vision-language pre-trained model (CLIP) to the unsupervised crowd counting task.
By fine-tuning the image encoder through multi-modal ranking loss and using a progressive filtering strategy, the performance of CrowdCLIP is largely improved. 
We conduct extensive experiments on challenging datasets to demonstrate the state-of-the-art performance of our CrowdCLIP. We hope our method can provide a new perspective for the crowd counting task.

\textbf{Acknowledgements.}
This work was supported by the National Science Fund for Distinguished Young Scholars of China (Grant No.62225603) and the Young Scientists Fund of the National Natural Science Foundation of China (Grant No.62206103). 


{\small
\bibliographystyle{ieee_fullname}
\bibliography{egbib}
}

\clearpage

\begin{appendices}

\setcounter{table}{0} 
\setcounter{figure}{0} 
\setcounter{equation}{0} 

\section{Inference speed}

We provide the comparison of inference speed, as shown in Tab.~\ref{tab:runtime}.
Note that the run time of our CrowdCLIP is dynamic, as the proposed progressive filtering strategy aims to choose high-confidence crowd patches, while the number of selected crowd patches from different images is different. To this end, we report the range of inference speed, \ie, [$24.0$, $50.8$], where the former means all patches of a given image contain human heads, and the latter means there are no crowd patches. Specifically, the fully supervised methods~\cite{li2018csrnet,ma2019bayesian} require high-resolution features to generate high-quality density maps, leading to low inference speed. Compared with the unsupervised SOTA CSS-CCNN~\cite{babu2022completely}, our method is highly competitive in terms of inference speed.

\begin{table}[htbp]
\small
\setlength{\tabcolsep}{3.5mm}
\caption{The comparisons of Frames Per Second (FPS) between our method and other methods. The results are conducted on an NVIDIA 3090 GPU.}
\centering

\begin{tabular}{ lccccc }
\toprule
Method & Label&Resolution  &FPS \\
 \midrule
CSRNet~\cite{li2018csrnet} &Point& $1024$ $\times$ $768$ & $18.4$ \\
BL~\cite{ma2019bayesian}&Point& $1024$ $\times$ $768$  &$21.3$ \\
\midrule
CSS-CCNN~\cite{babu2022completely} &\textcolor{red}{None}&$1024$ $\times$ $768$  & $37.4$ \\
\textbf{CrowdCLIP} &\textcolor{red}{None} & $1024$ $\times$ $768$  & [$24.0$, $50.8$] \\
 \bottomrule
\end{tabular}
\label{tab:runtime}
\end{table}


\section{More visualizations}

\begin{figure*}[htbp]
	\begin{center}
		\includegraphics[width=0.96\linewidth]{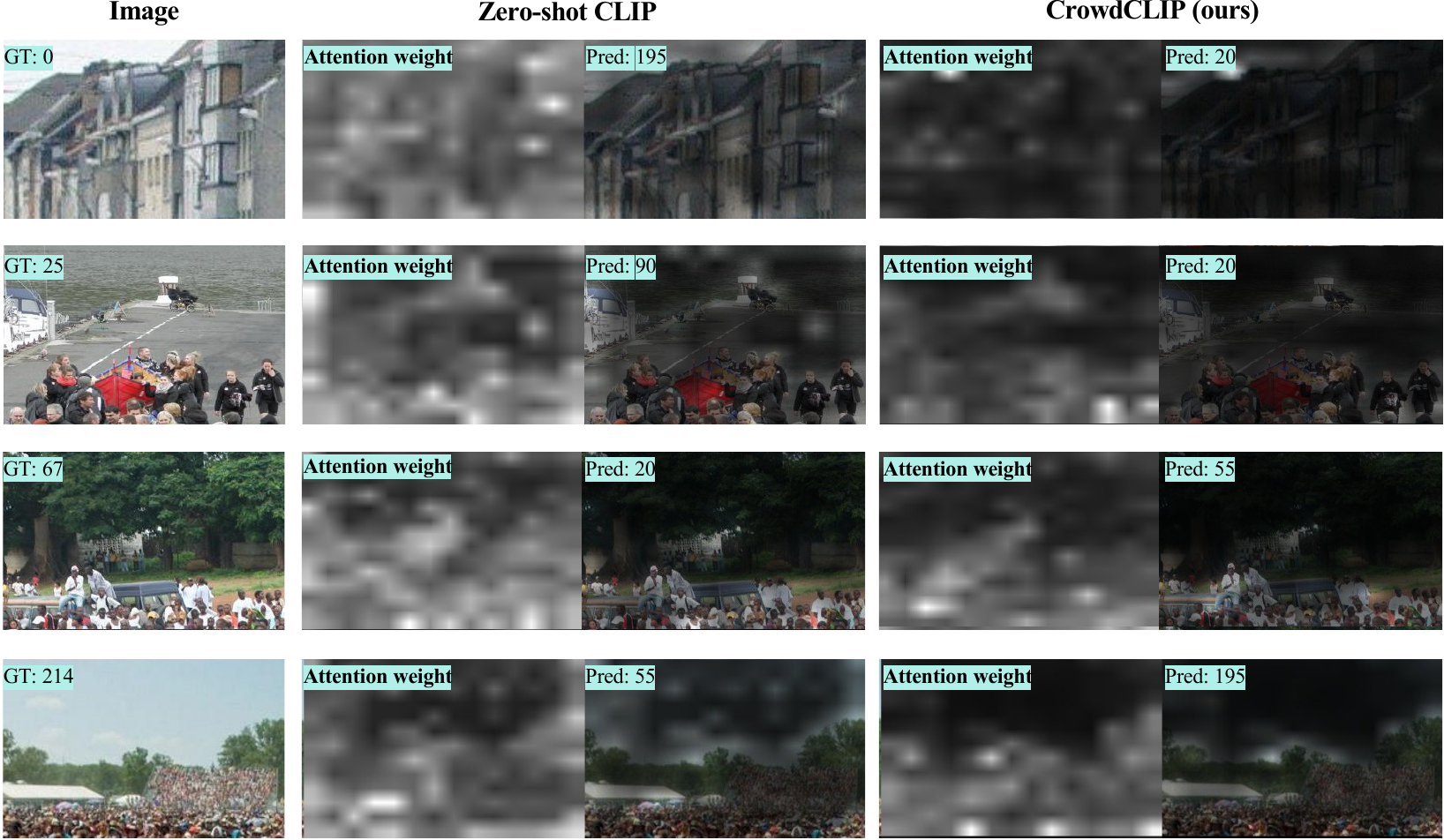}
	\end{center}
	\caption{
	Qualitative visualizations of zero-shot CLIP and our proposed CrowdCLIP. From left to right, there are ground truth, results from zero-shot CLIP, and results from CrowdCLIP. 
 }
	\label{fig:supple_vis}
\end{figure*}

\textbf{Qualitative comparisons.} We provide qualitative comparisons to further demonstrate the effectiveness of our method, as shown in Fig.~\ref{fig:supple_vis}. Specifically, we can observe that the zero-shot CLIP\footnote{Zero-shot CLIP means directly adopting the original non-fine-tuned image encoder of CLIP.} can not understand crowd semantics well, leading to poor performance. In contrast, the proposed CrowdCLIP can generate more reasonable attention through the proposed ranking-based contrastive fine-tuning, resulting in better performance. 

\begin{figure*}[t]
	\begin{center}
		\includegraphics[width=0.96\linewidth]{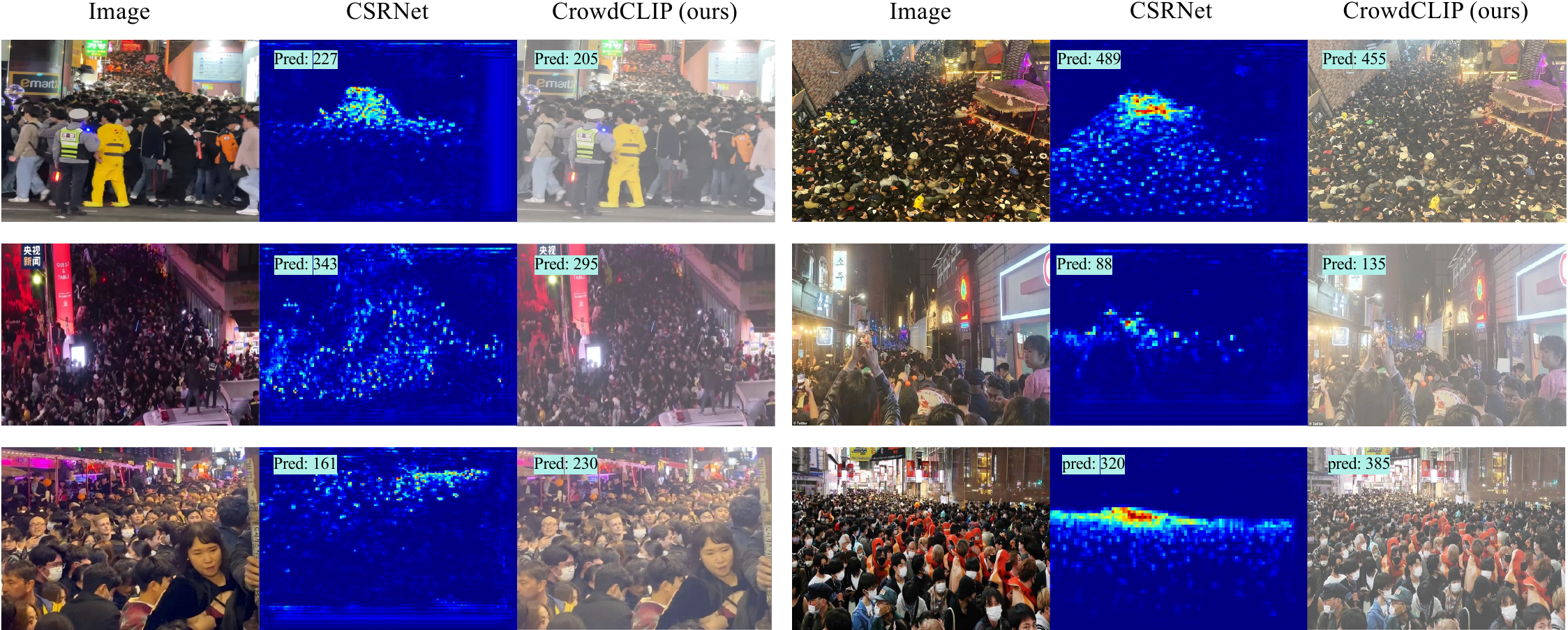}
	\end{center}
	\caption{
Evaluations of CSRNet~\cite{li2018csrnet} and the proposed CrowdCLIP on the Seoul Halloween crowd crush scenes. The two models are trained on the UCF-QNRF dataset.
 }
	\label{fig:litai}
\end{figure*}

\textbf{Testing on Seoul Halloween crowd crush scenes.} On the night of $29$ October $2022$, a crowd crush occurred during Halloween festivities in the Itaewon neighborhood of Seoul, South Korea. At least $158$ people were killed, and $196$ others were injured. One of the reasons is that hundreds of people simultaneously appear in the narrow alley. A good crowd counting algorithm will help relieve the crowd crush event. In this part, we test the CrowdCLIP on the Seoul Halloween crowd crush scenes, as shown in Fig.~\ref{fig:litai}. Note that there is no ground truth for these images, so we choose a representative fully-supervised counting method CSRNet~\cite{li2018csrnet} as a comparison. We can find that the prediction of our method is close to the CSRNet in most cases. 

\end{appendices}

\end{document}